\documentclass{article}

\usepackage{PRIMEarxiv}

\usepackage[utf8]{inputenc} 
\usepackage[T1]{fontenc}    
\usepackage{url}            
\usepackage{booktabs}       
\usepackage{amsfonts}       
\usepackage{nicefrac}       
\usepackage{microtype}      
\usepackage{lipsum}
\usepackage{fancyhdr}       
\usepackage{graphicx}       
\graphicspath{{media/}}     

\usepackage{hyperref}
\hypersetup{colorlinks,allcolors=black}

\usepackage{times}
\usepackage{latexsym}

\usepackage[T1]{fontenc}

\usepackage[utf8]{inputenc}

\usepackage{microtype}

\usepackage{inconsolata}
\usepackage{microtype}
\usepackage{graphicx}

\usepackage[textsize=tiny]{todonotes}
\usepackage{amssymb}

\usepackage{adjustbox}
\usepackage{graphicx}
\usepackage{caption}
\usepackage{subcaption}

\usepackage{amsmath}
\usepackage{bm}
\usepackage{multirow}
\usepackage{xcolor}
\usepackage[utf8]{inputenc}
\usepackage{mathtools}
\usepackage{bbm}
\usepackage{physics}
\usepackage{float}
\usepackage{booktabs}
 \usepackage{array, makecell}
 \usepackage{algorithm}
 \usepackage{algorithmic}
\usepackage{tabularx}
\usepackage{amsthm}

\title{Does Faithfulness Conflict with Plausibility? An Empirical Study in Explainable AI across NLP Tasks
}

\author{
  Xiaolei Lu\\
  \texttt{xiaoleilu2-c@my.cityu.edu.hk} \\
   \And
  Jianghong Ma \\
  \texttt{majianghong@hit.edu.cn} \\
}

\begin{document}
\maketitle

\begin{abstract}
Explainability algorithms aimed at interpreting decision-making AI systems usually consider balancing two critical dimensions: 1) \textit{faithfulness}, where explanations accurately reflect the model's inference process. 2) \textit{plausibility}, where explanations are consistent with domain experts. However, the question arises: do faithfulness and plausibility inherently conflict? In this study, through a comprehensive quantitative comparison between the explanations from the selected explainability methods and expert-level interpretations across three NLP tasks: sentiment analysis, intent detection, and topic labeling, we demonstrate that traditional perturbation-based methods Shapley value and LIME could attain greater faithfulness and plausibility. Our findings suggest that rather than optimizing for one dimension at the expense of the other, we could seek to optimize explainability algorithms with dual objectives to achieve high levels of accuracy and user accessibility in their explanations.
\end{abstract}

\keywords{Explainability \and Faithfulness\and Plausibility}

\section{Introduction}
Deep Neural Networks (DNNs) have demonstrated impressive results in many domains including Natural Language Processing (NLP), Computer Vision (CV) and speech processing \cite{devlin2018bert,achiam2023gpt}. These deep neural models operate like black-box models by applying multiple layers of non-linear transformation on the vector representations of input data, which fails to provide insights to understand the inference process.

Explainability algorithms including attention-based, gradient-based and perturbation-based feature attribution methods have been extensively studied to explore the internal mechanisms of black-box deep models \cite{wiegreffe2019attention,sundararajan2017axiomatic,shapley1953value}, which improves transparency in AI, particularly in sensitive applications like clinical decision-making systems. A good explanation in such contexts should consider two critical dimensions: 1) \textit{Faithfulness}. The explanation could accurately attribute the model's decision to the specific features. 2) \textit{Plausibility}. The explanation is logically sound and understandable to the domain experts. 

As plausibility focuses on the human's perception of the explanation, more faithful explanations that accurately convey the reasoning of complex models (e.g. deep neural networks) may be implausible to a domain expert, and vice versa. Explainability research commonly recognizes a trade-off between faithfulness and plausibility, suggesting that enhancing one may compromise the other \cite{wood2021faithful,jacovi2020towards}. However, few studies explicitly address the conflicts between these dimensions during evaluation, which requires further empirical investigations. 

In this work, through comprehensive quantitative analysis we evaluate the explanations from the selected explainability methods and expert-level interpretations across NLP tasks. Our contributions are summarized as follows:
\begin{itemize}
\item We utilize GPT-4, which demonstrates its expert role in our consistency verification, to construct professional explanations across our targeted datasets, serving as benchmarks for plausibility evaluation. 
\item We thoroughly evaluate the faithfulness and plausibility of  explanations from GPT-4 and the selected explainability methods. Our findings suggest the possibility of optimizing explainability algorithms to simultaneously achieve high performance in both faithfulness and plausibility.
\end{itemize}

\section{Related Work}

Existing explainability methods for interpreting model training and inference process could be categorized into two types: instance attribution measures how a training point influence the prediction of a given instance while feature attribution quantifies the contribution of each feature (or feature interaction) to the model's output on a specific instance. For example, Influence Function \cite{koh2017understanding} attends to the final iteration of the training and computes the influence score of a training instance on the prediction loss of a test sample. Shapley value \cite{shapley1953value} that is derived from cooperative game theory treats each feature as a player and computes the marginal contribution of each feature toward the model's output. Integrated Gradients \cite{sundararajan2017axiomatic} measures feature importance by computing the path integral of the gradients respect to each dimension of input. 

Faithfulness and plausibility are the primary criteria to evaluate explainability methods. Previous works \cite{el2022evaluation, sato2022plausibility} collected ground truth explanations (or rationales) from crowdworkers with human-to-human agreement. However, plausible explanations may not be faithful to reflect the model's inference process. Proxy model \cite{wood2021faithful} was proposed to use the trained
model’s predictions as training labels to balance faithfulness and plausibility. Since obtaining professional human explanations is challenging, few studies explicitly address the conflicts between faithfulness and plausibility.

\section{Experimental Setup}

\subsection{Tasks, Datasets and Models}

We conduct experiments on various NLP tasks including sentiment analysis, intent detection and topic labeling. The employed datasets are SST-2 \cite{socher2013recursive}, SNIPS \cite{coucke2018snips} and 20Newsgroups\footnote{http://qwone.com/~jason/20Newsgroups/}, and study the performance on BERT-base \cite{devlin2018bert} and RoBERTa-base \cite{liu2019roberta} models. Appendix \ref{appendixa} provides the configurations of finetuning pretrained BERT-base and RoBERTa-base models in these downstream tasks. To enable user-friendly human evaluation, we select the explained set where the sequence remains unchanged after tokenization. Appendix \ref{appendixb} summarizes the details of the datasets.

\subsection{Explanation Methods}
We study model explainability from three groups of explanation methods: attention-based, gradient-based and perturbation-based attributions. The employed attribution methods are described as follows: \textbf{Inherent Attention Explanation (RawAtt)}\cite{wiegreffe2019attention}: directly measure the feature importance with attention weights. \textbf{Attention Rollout (AttRll)} \cite{abnar2020quantifying}: aggregate attention across all heads and layers to measure how much each input feature attends to every other feature across the entire depth of the model. \textbf{Input $\odot$ Gradients (InputG)} \cite{shrikumar2017learning}: measure the change of the model's output with respect to a small change in the input feature. \textbf{Integrated Gradients (IG)} \cite{sundararajan2017axiomatic}: accumulate the gradients along the path from a given baseline to the input. \textbf{Shapley value (SV)} \cite{shapley1953value}: average marginal contribution of the feature being explained toward the model's output over all possible permutations. \textbf{LIME} \cite{ribeiro2016should}: generate explanation by learning an inherently interpretable model locally on the instance being explained.

\begin{figure*}[h]
 \centering
 \begin{subfigure}[b]{\textwidth}
     \centering
     \includegraphics[width=\textwidth,height=1in]{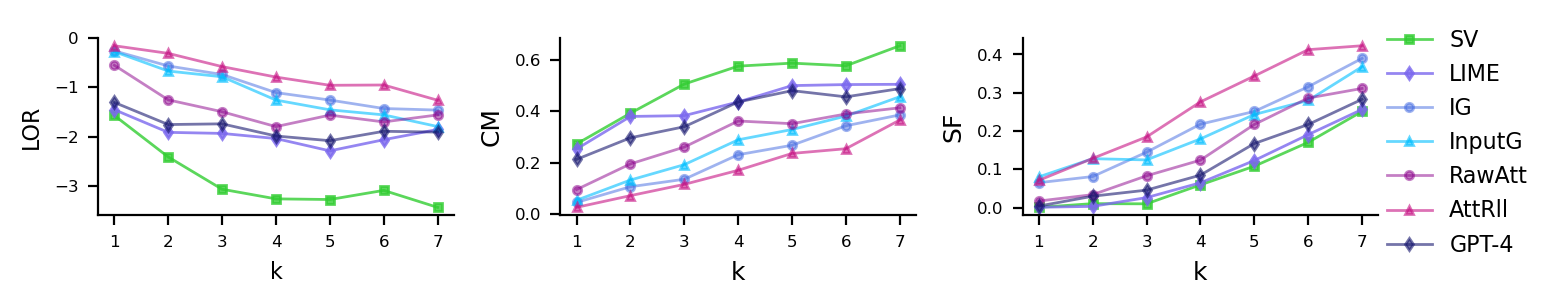}
     \caption{Faithfulness evaluation performance over BERT on SST-2.}
     \label{fig1}
 \end{subfigure}
 \begin{subfigure}[b]{\textwidth}
     \centering
     \includegraphics[width=\textwidth,height=0.9in]{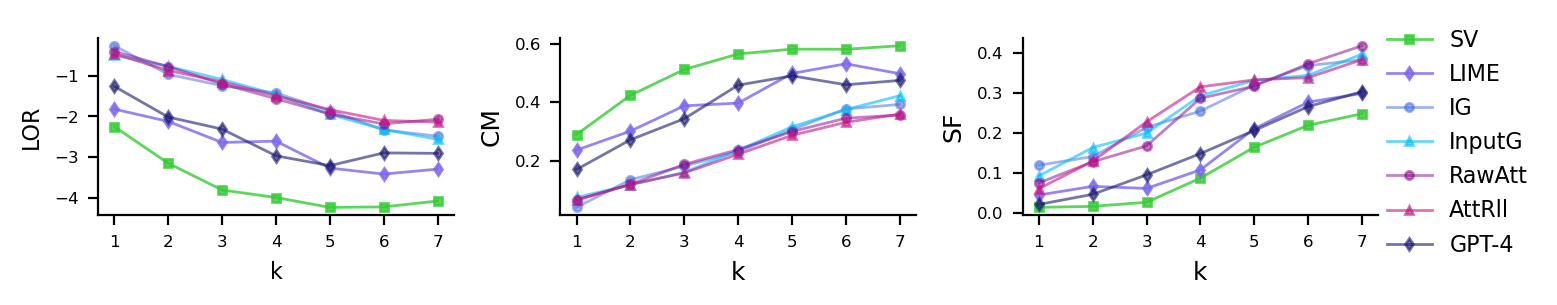}
     \caption{Faithfulness evaluation performance over RoBERTa on SST-2.}
     \label{fig2}
 \end{subfigure}
 \begin{subfigure}[b]{\textwidth}
     \centering
     \includegraphics[width=\textwidth,height=0.9in]{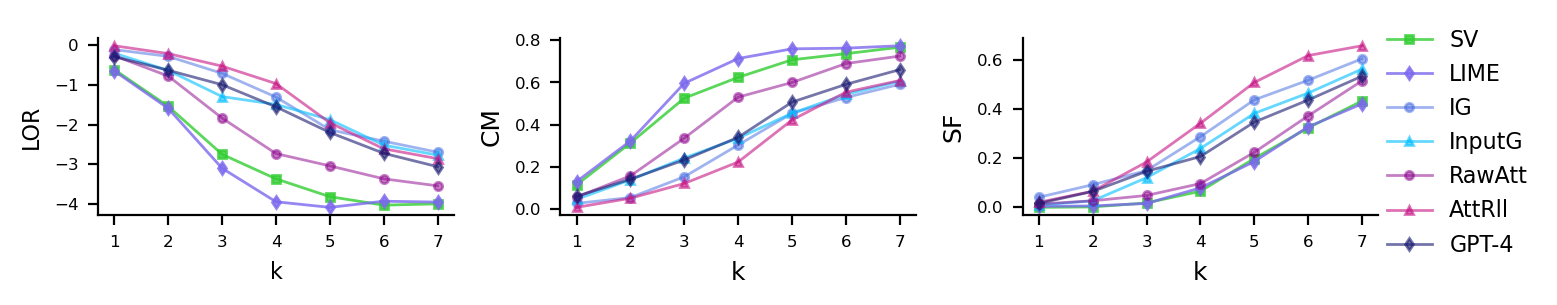}
     \caption{Faithfulness evaluation performance over BERT on SNIPS.}
     \label{fig3}
 \end{subfigure}
 \begin{subfigure}[b]{\textwidth}
     \centering
     \includegraphics[width=\textwidth,height=0.9in]{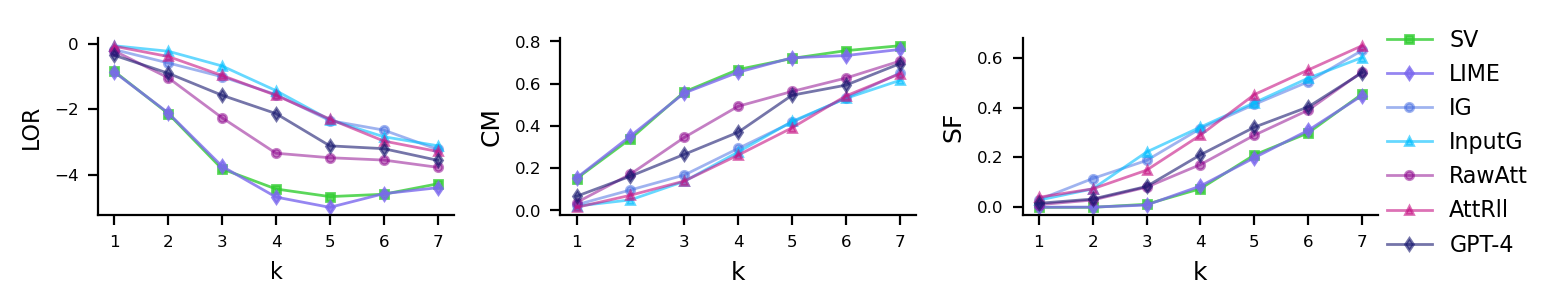}
     \caption{Faithfulness evaluation performance over RoBERTa on SNIPS.}
     \label{fig4}
 \end{subfigure}
 
 \caption{Faithfulness evaluation performances on SST-2 and SNIPS over BERT and RoBERTa architectures, where lower LOR and SF scores are better, higher CM scores are preferred. }
 \label{er}
\end{figure*}

\subsection{Evaluation matrices}

\textbf{Faithfulness}: three faithful evaluation metrics are employed and we choose padding replacement operation\footnote{Deletion operation produces the similar results over BERT architecture. RoBERTa's training process is more robust to variations in input and padding and deletion operations generate the same results over RoBERTa architecture.}. The employed matrices include \textbf{Log-odds (LOR)} \cite{shrikumar2017learning}: average the difference of negative logarithmic probabilities on the predicted class over the test
data before and after replacing the top $k$ influential words from the text sequence. \textbf{Sufficiency (SF)} \cite{deyoung2019eraser}: measure whether important features identified by the explanation method are adequate to remain confidence on the original predictions. \textbf{Comprehensiveness (CM)} \cite{deyoung2019eraser}: evaluate if the features assigned lower weights are unnecessary for the predictions. 

\textbf{Plausibility}: evaluate the similarity between the feature importance ranking generated by the explanation methods and GPT-4. Given the $i_{th}$ input sequence with size $n$, let $H_i$ denotes the human explanation toward feature importance ranking and $E_i$ is provided by the explainability method. The employed matrices include \textbf{Rank Correlation (RC)}: measure the similarity between two ranks. Spearman’s Rank Correlation Coefficient is employed to compute RC. \textbf{Overlap Rate (OR)}: measure the overlap of the top $k$ influential elements between $E_{i,k}$ and $H_{i,k}$.

\subsection{Human explanation}
There are some research reports \cite{tornberg2023chatgpt,feng2023investigating} showing that large language models (LLMs), such as GPT-3.5 and GPT-4, can provide high-quality annotations like an excellent crowdsourced annotator does. We first randomly select 83 explained instances for BERT and 74 for RoBERTa on SST-2, 86 explained instances for both BERT and RoBERTa on SNIPS. By comparing the explanations generated by GPT-4 (details of the prompt for generating explanation is given in Appendix \ref{appendixc}) and an NLP researcher, Rank Correlation scores are 0.71 and 0.77 for BERT and RoBERTa, respectively, on SST-2, and 0.86 and 0.83, respectively, on SNIPS. These results demonstrate the quality of GPT-4 in the expert role. Therefore we use GPT-4 to provide explanations toward the model's output for more instances (we will provide the explained sets with corresponding GPT-4 explanations including our consistency verification in the public version).

\begin{table*}
\centering
\begin{tabular}{lcccccccc} 
\toprule
\multicolumn{2}{c}{Method}     & SV      & LIME    & IG      & InputG  & RawAtt  & AttRll  & GPT-4    \\ 
\hline
\multirow{3}{*}{BERT}    & LOR & -5.9748 & -3.5052 & -0.9578 & -1.1743 & -2.2261 & -0.6265 & -3.2694  \\
                         & CM  & 0.8874  & 0.6880  & 0.2156  & 0.2677  & 0.4352  & 0.1330  & 0.5848   \\
                         & SF  & -0.0572 & 0.1360  & 0.6132  & 0.5600  & 0.4189  & 0.6815  & 0.3071   \\ 
\hline
\multirow{3}{*}{RoBERTa} & LOR & -5.4660 & -3.1295 & -1.2463 & -1.2423 & -0.9516 & -0.5808 & -3.4327  \\
                         & CM  & 0.8392  & 0.5721  & 0.2238  & 0.2315  & 0.2021  & 0.1217  & 0.5748   \\
                         & SF  & -0.0868 & 0.2523  & 0.5217  & 0.5160  & 0.6284  & 0.6451  & 0.3367   \\
\bottomrule
\end{tabular}
\caption{Faithfulness evaluation performances on 20Newsgroup over BERT and RoBERTa architectures.}
\label{err}
\end{table*}

\section{Results}
Fig.\ref{er} and Table \ref{err} demonstrate the faithfulness evaluation performance \footnote{For 20Newsgroup we use GPT-4 to provide the most positive influential features and the remaining treated as unimportant features.}. We can observe that in SST-2 and 20Newsgroup over BERT and RoBERTa architectures SV outperforms the other baselines, and LIME and GPT-4 obtain comparative performance which is second only to that of SV. In SNIPS, both LIME and SV achieve similar outcomes, whereas GPT-4's performance is moderate.

Generally, SV, LIME and GPT-4 outperform the selected gradient-based and attention-based methods in these datasets. First, attention-based methods assume higher attention weights correlate with higher importance while these weights may also contain additional information that could be utilized by the downstream models \cite{bai2021attentions}. Compared with the perturbation-based attribution that is less reliant on the model architecture, gradient-based methods might not accurately measure how input features affect the output of the complex non-linear model. Furthermore, the plausible explanations provided by the expert (e.g. GPT-4) could be more faithful than some explainability algorithms. 

\begin{figure}
 \centering
 \begin{subfigure}[b]{0.4\textwidth}
     \centering
     \includegraphics[width=\textwidth,height=1.5in]{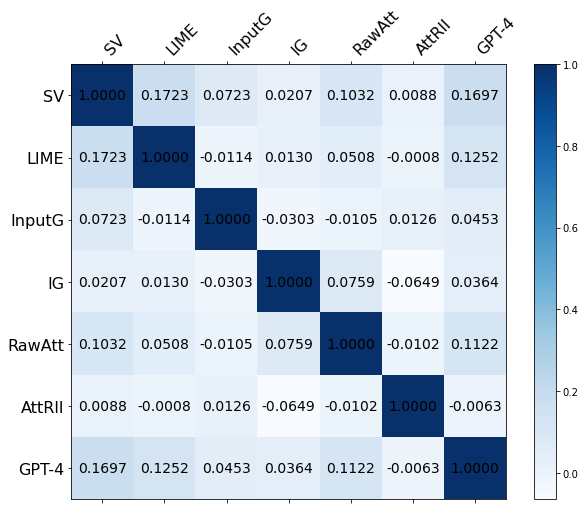}
     \caption{SST-2 over BERT.}
     \label{si1}
 \end{subfigure}
 \begin{subfigure}[b]{0.4\textwidth}
     \centering
     \includegraphics[width=\textwidth,height=1.5in]{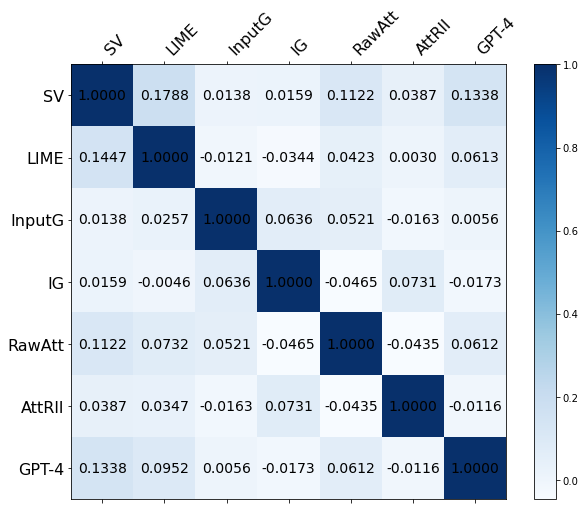}
     \caption{SNIPS over BERT.}
     \label{ag1}
 \end{subfigure}
  \begin{subfigure}[b]{0.4\textwidth}
     \centering
     \includegraphics[width=\textwidth,height=1.5in]{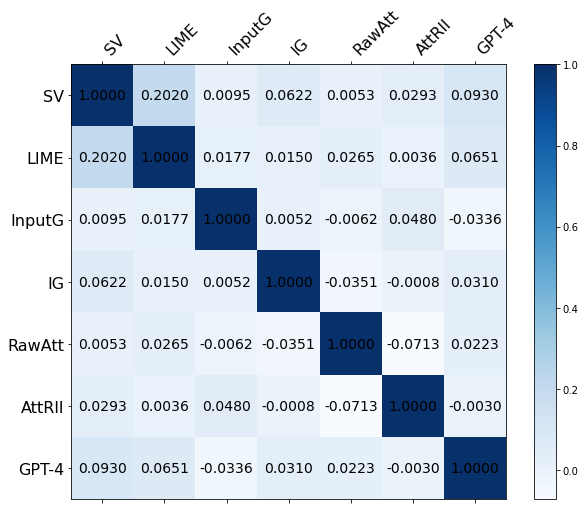}
     \caption{SST-2 over RoBERTa.}
     \label{si2}
 \end{subfigure}
 \begin{subfigure}[b]{0.4\textwidth}
     \centering
     \includegraphics[width=\textwidth,,height=1.5in]{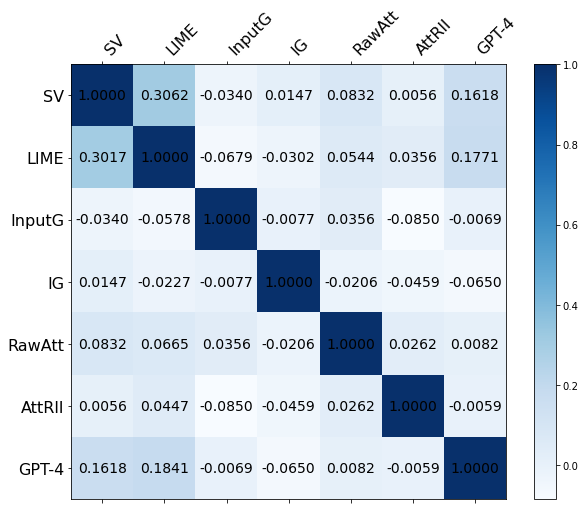}
     \caption{SNIPS over RoBERTa.}
     \label{ag2}
 \end{subfigure}
 \caption{Rank coefficient between the explainability methods and GPT-4 on SST-2 and SNIPS over BERT and RoBERTa architectures.}
 \label{rc2}
\end{figure}

Fig.\ref{rc2} presents the rank coefficient between the explainability methods and GPT-4 on SST-2 and SNIPS over BERT and RoBERTa architectures. Overall the explanations toward feature importance ranking between the explainability methods and GPT-4 are weakly correlated. We further examine the overlap rate of the explanations between these methods. Appendix \ref{appendixe} Fig.\ref{op4} shows the overlap rate of the feature importance ranking ($k=4$) between the explainability methods and GPT-4. SV and LIME achieve over $60\%$ OR with GPT-4 in identifying the most positive influential features in the SST-2 and SNIPS datasets, and these two methods also obtain higher OR values in 20Newsgroup. Despite the weak correlation when considering the full importance ranking, significant overlap exists in identifying the most critical features between the chosen methods and GPT-4. We also present the overlap rate with different $k$ in Appendix \ref{appendixe}.

\section{Conclusion}
In this work by conducting experiments on three NLP tasks and constructing expert-level human explanations with GPT-4, we quantitatively analyze the explanations from the selected explainability methods and human-generated interpretations toward NLP deep models. The results show that SV, LIME, and GPT-4 outperform traditional gradient-based and attention-based methods across various datasets. Our findings suggest that plausibility and faithfulness can be complementary. The explainability method could achieve a high overlap rate in identifying influential features and also tend to provide explanations that are plausible to human interpreters, which implies that the explainability algorithms can be optimized toward the dual objective of faithfulness and plausibility.

\section{Limitations}

This empirical study focuses on a selected set of explainability methods across three specific NLP tasks: sentiment analysis, intent detection, and topic labeling. While the results provide insights into the relationship between faithfulness and plausibility, it also limits the generalizability of our findings. Future research could include more tasks and models. Furthermore, our findings suggest that explainability algorithms could be optimized to achieve both faithfulness and plausibility. How to optimize these algorithms for multiple objectives simultaneously require further investigation.

\bibliographystyle{unsrt}  
\bibliography{reference}  

\begin{thebibliography}{10}

\bibitem{devlin2018bert}
Jacob Devlin, Ming-Wei Chang, Kenton Lee, and Kristina Toutanova.
\newblock Bert: Pre-training of deep bidirectional transformers for language understanding.
\newblock {\em arXiv preprint arXiv:1810.04805}, 2018.

\bibitem{achiam2023gpt}
Josh Achiam, Steven Adler, Sandhini Agarwal, Lama Ahmad, Ilge Akkaya, Florencia~Leoni Aleman, Diogo Almeida, Janko Altenschmidt, Sam Altman, Shyamal Anadkat, et~al.
\newblock Gpt-4 technical report.
\newblock {\em arXiv preprint arXiv:2303.08774}, 2023.

\bibitem{wiegreffe2019attention}
Sarah Wiegreffe and Yuval Pinter.
\newblock Attention is not not explanation.
\newblock {\em arXiv preprint arXiv:1908.04626}, 2019.

\bibitem{sundararajan2017axiomatic}
Mukund Sundararajan, Ankur Taly, and Qiqi Yan.
\newblock Axiomatic attribution for deep networks.
\newblock In {\em International conference on machine learning}, pages 3319--3328. PMLR, 2017.

\bibitem{shapley1953value}
Lloyd~S Shapley et~al.
\newblock A value for n-person games.
\newblock 1953.

\bibitem{wood2021faithful}
Zach Wood-Doughty, Isabel Cachola, and Mark Dredze.
\newblock Faithful and plausible explanations of medical code predictions.
\newblock {\em arXiv preprint arXiv:2104.07894}, 2021.

\bibitem{jacovi2020towards}
Alon Jacovi and Yoav Goldberg.
\newblock Towards faithfully interpretable nlp systems: How should we define and evaluate faithfulness?
\newblock {\em arXiv preprint arXiv:2004.03685}, 2020.

\bibitem{koh2017understanding}
Pang~Wei Koh and Percy Liang.
\newblock Understanding black-box predictions via influence functions.
\newblock In {\em International conference on machine learning}, pages 1885--1894. PMLR, 2017.

\bibitem{el2022evaluation}
Julia El~Zini, Mohamad Mansour, Basel Mousi, and Mariette Awad.
\newblock On the evaluation of the plausibility and faithfulness of sentiment analysis explanations.
\newblock In {\em IFIP International Conference on Artificial Intelligence Applications and Innovations}, pages 338--349. Springer, 2022.

\bibitem{sato2022plausibility}
Tasuku Sato, Hiroaki Funayama, Kazuaki Hanawa, and Kentaro Inui.
\newblock Plausibility and faithfulness of feature attribution-based explanations in automated short answer scoring.
\newblock In {\em International Conference on Artificial Intelligence in Education}, pages 231--242. Springer, 2022.

\bibitem{socher2013recursive}
Richard Socher, Alex Perelygin, Jean Wu, Jason Chuang, Christopher~D Manning, Andrew~Y Ng, and Christopher Potts.
\newblock Recursive deep models for semantic compositionality over a sentiment treebank.
\newblock In {\em Proceedings of the 2013 conference on empirical methods in natural language processing}, pages 1631--1642, 2013.

\bibitem{coucke2018snips}
Alice Coucke, Alaa Saade, Adrien Ball, Th{\'e}odore Bluche, Alexandre Caulier, David Leroy, Cl{\'e}ment Doumouro, Thibault Gisselbrecht, Francesco Caltagirone, Thibaut Lavril, et~al.
\newblock Snips voice platform: an embedded spoken language understanding system for private-by-design voice interfaces.
\newblock {\em arXiv preprint arXiv:1805.10190}, 2018.

\bibitem{liu2019roberta}
Yinhan Liu, Myle Ott, Naman Goyal, Jingfei Du, Mandar Joshi, Danqi Chen, Omer Levy, Mike Lewis, Luke Zettlemoyer, and Veselin Stoyanov.
\newblock Roberta: A robustly optimized bert pretraining approach.
\newblock {\em arXiv preprint arXiv:1907.11692}, 2019.

\bibitem{abnar2020quantifying}
Samira Abnar and Willem Zuidema.
\newblock Quantifying attention flow in transformers.
\newblock {\em arXiv preprint arXiv:2005.00928}, 2020.

\bibitem{shrikumar2017learning}
Avanti Shrikumar, Peyton Greenside, and Anshul Kundaje.
\newblock Learning important features through propagating activation differences.
\newblock In {\em International conference on machine learning}, pages 3145--3153. PMLR, 2017.

\bibitem{ribeiro2016should}
Marco~Tulio Ribeiro, Sameer Singh, and Carlos Guestrin.
\newblock " why should i trust you?" explaining the predictions of any classifier.
\newblock In {\em Proceedings of the 22nd ACM SIGKDD international conference on knowledge discovery and data mining}, pages 1135--1144, 2016.

\bibitem{deyoung2019eraser}
Jay DeYoung, Sarthak Jain, Nazneen~Fatema Rajani, Eric Lehman, Caiming Xiong, Richard Socher, and Byron~C Wallace.
\newblock Eraser: A benchmark to evaluate rationalized nlp models.
\newblock {\em arXiv preprint arXiv:1911.03429}, 2019.

\bibitem{tornberg2023chatgpt}
Petter T{\"o}rnberg.
\newblock Chatgpt-4 outperforms experts and crowd workers in annotating political twitter messages with zero-shot learning.
\newblock {\em arXiv preprint arXiv:2304.06588}, 2023.

\bibitem{feng2023investigating}
Yunhe Feng, Sreecharan Vanam, Manasa Cherukupally, Weijian Zheng, Meikang Qiu, and Haihua Chen.
\newblock Investigating code generation performance of chatgpt with crowdsourcing social data.
\newblock In {\em Proceedings of the 47th IEEE Computer Software and Applications Conference}, pages 1--10, 2023.

\bibitem{bai2021attentions}
Bing Bai, Jian Liang, Guanhua Zhang, Hao Li, Kun Bai, and Fei Wang.
\newblock Why attentions may not be interpretable?
\newblock In {\em Proceedings of the 27th ACM SIGKDD Conference on Knowledge Discovery \& Data Mining}, pages 25--34, 2021.

\end{thebibliography}

\appendix

\section{Configurations for finetuning deep models}
\label{appendixa}
We use AdamW optimizer with weight decay $0.001$ and start with learning rate of 2e-5 to tune pretrained BERT-base-uncased and RoBERTa-base models. For the setting of epochs and batch size, SST-2 at 10 epochs with a batch size of 32, SNIPS at 10/64, and NG at 20/64, ensuring good model performance for each task. The corresponding performance is reported in Table \ref{appendixff}.

\begin{table}[H]
\centering
\begin{tabular}{lccc} 
\toprule
Models  & SST-2   & SNIPS & 20Newsgroup  \\ 
\hline
BERT    & 90.49  & 97.71 & 74.48        \\
RoBERTa & 94.56  & 97.85 & 73.37        \\
\bottomrule
\end{tabular}
\caption{Task performance (\%) with finetuning BERT and RoBERTa.}
\label{appendixff}
\end{table}

\section{Details of the tasks and datasets}
\label{appendixb}
Table \ref{dasum} summarizes the selected datasets with corresponding explained sets.

\renewcommand{\arraystretch}{1.5}
\begin{table*}
\centering
\begin{tabular}{lccccccc} 
\toprule
\multirow{2}{*}{Datasets} & \multirow{2}{*}{Train set} & \multirow{2}{*}{Test set} & \multirow{2}{*}{Label set} & \multicolumn{2}{c}{BERT}                                               & \multicolumn{2}{c}{RoBERTa}                                             \\
                          &                            &                           &                            & \multicolumn{1}{l}{Explained set} & \multicolumn{1}{l}{Avg\_len} & \multicolumn{1}{l}{Explained set} & \multicolumn{1}{l}{Avg\_len}  \\ 
\hline
SST-2                     & 6899                       & 1819                      & 2                          & 152                               & 7.39                               & 164                               & 8.88                                \\
SNIPS                     & 13082                      & 700                       & 7                          & 188                               & 7.35                               & 194                               & 7.50                                \\
20Newsgroup               & 10663                      & 7019                      & 20                         & 89                                & 23.66                              & 78                                & 29.03                               \\
\bottomrule
\end{tabular}
\caption{Summary of the selected datasets, where Avg\_len denotes the average length.}
\label{dasum}
\end{table*}

\section{Prompt for GPT-4 to generate explanations}
\label{appendixc}

In this section, we demonstrate how to use GPT-4 to generate explanations via Table \ref{t1}, \ref{t2} and \ref{t3}. ``Input'' and ``Output'' refer to the prompt provided to GPT-4 and the generated explanations, respectively. It could be treated as zero-shot evaluation. We maintained the output integrity without alternations, while occasionally adjusting the requirements to ensure a complete ranking. For instance, when dealing with repeated strings, each instance was assigned an individual rank. Due to the lengthy context in the 20Newsgroup dataset, we only use GPT-4 to provide the most positive influential features toward the model's output.

\begin{table*}[]
\centering
\begin{tabularx}{\textwidth}{|c|X|}
\hline
\textbf{Component} & \textbf{Description} \\ \hline
\multirow{8}{*}{Input} & The task is described as follows: given a text sequence of a movie review with the sentiment classification label (positive or negative), there are a few requirements:
\begin{enumerate}
    \item Transform this long string sequence into a list of strings.
    \item Measure the contributions of each string in the list toward the sentiment label based on your understanding. Then rank all strings (ensuring that no strings are excluded) including the repeated strings (each occurrence should have its own rank) in this list based on their contributions.
    \item The ranking should follow an order from the most positive to neutral to the most negative. Place the strings with the highest positive contribution at the top and the strings with the most negative contribution at the bottom.
    \item Output all ranked strings ensuring that no strings are excluded.
\end{enumerate} \\ \hline
\multirow{2}{*}{Example Input} & Sequence: something the true film buff will enjoy \newline
Label: positive \\ \hline
\multirow{1}{*}{Example Output} & 
Ranked strings: [`enjoy', `true', `something', `film', `buff', `will', `the'] \\ \hline
\end{tabularx}
\caption{Illustration of how to use GPT-4 to generate explanations for SST-2 explained set.}
\label{t1}
\end{table*}

\begin{table*}[]
\centering
\begin{tabularx}{\textwidth}{|c|X|}
\hline
\textbf{Component} & \textbf{Description} \\ \hline
\multirow{8}{*}{Input} & The task is described as follows: given the utterance with the corresponding predicted intent, by treating yourself as an human, please rank all words based on their influence toward this predicted intent. 
\begin{enumerate}
    \item Transform this long string sequence into a list of strings.
    \item Treat yourself as a human and rank each individual string (ensuring that no strings are excluded), including the repeated strings, in this list based on their contributions toward the predicted intent.
    \item The ranking should follow an order from the most positive to neutral to the most negative. Place the strings with the highest positive contribution at the top and the strings with the most negative contribution at the bottom.
    \item Output all ranked strings ensuring that no strings are excluded.
\end{enumerate} \\ \hline
\multirow{2}{*}{Example Input} & Sequence: find an album called just call me stupid \newline
Label: search creative work \\ \hline
\multirow{1}{*}{Example Output} & Ranked strings: [`find', `album', `called', `just', `call', `stupid', `me', `an']
 \\ \hline
\end{tabularx}
\caption{Illustration of how to use GPT-4 to generate explanations for SNIPS explained set.}
\label{t2}
\end{table*}

\begin{table*}[]
\centering
\begin{tabularx}{\textwidth}{|c|X|}
\hline
\textbf{Component} & \textbf{Description} \\ \hline
\multirow{8}{*}{Input} &  The task is described as follows: given a news with the corresponding topic, please provide the evaluation for the topic labeling of this news. There are a few specified requirements:
\begin{enumerate}

\item Transform this long string sequence to a list of strings.
\item Treat yourself as a human and find out the most positive influential strings toward the predicted topic.
\item Output the ranked list.
\end{enumerate} \\ \hline
\multirow{2}{*}{Example Input} & Sequence: does anyone know where i can get some voice synthesis chips i am looking for something like the ones that do the time and date stamp on answering machines \newline
Topic: discussions about electronics \\ \hline
\multirow{1}{*}{Example Output} & 
Ranked strings: [`voice', `synthesis', `chips', `time', `date', `stamp']
\\ \hline
\end{tabularx}
\caption{Illustration of how to use GPT-4 to generate explanations for 20Newsgroup explained set.}
\label{t3}
\end{table*}

\section{Evaluation matrices}

\textbf{Log-odds (LOR)} \cite{shrikumar2017learning}: average the difference of negative logarithmic probabilities on the predicted class over the test
data before and after replacing the top $k$ influential words from the text sequence.

\begin{equation}
    \mathrm{LOR}(k) = \frac{1}{N}\sum_{i=1}^N \log \frac{f(\bm{x'}_i)}{f(\bm{x}_i)},
\end{equation}
where $\bm{x'}_i$ is obtained by replacing the $k$ top-scored words from $\bm{x}_i$. The lower LOR, the more faithful feature importance ranking.

\textbf{Sufficiency (SF)} \cite{deyoung2019eraser}: measure whether important features identified by the explanation method are adequate to remain confidence on the original predictions. 
\begin{equation}
    \mathrm{SF}(k) = \frac{1}{N}\sum_{i=1}^N f(\bm{x}_i)-f(\bm{x}_{i,k}),
\end{equation}
where $\bm{x}_{i,k}$ is obtained by replacing non-top $k$ influential elements in $\bm{x}_i$. The lower SF, the more faithful feature importance ranking.

\textbf{Comprehensiveness (CM)} \cite{deyoung2019eraser}: evaluate if the features assigned lower weights are unnecessary for the predictions.
\begin{equation}
   \mathrm{CM}(k) = \frac{1}{N}\sum_{i=1}^N f(\bm{x}_i)-f(\bm{x}_{i}\backslash \bm{x}_{i,k}),
\end{equation}
where $\bm{x}_{i}\backslash \bm{x}_{i,k}$ is obtained by replacing top $k$ influential elements in $\bm{x}_i$. The higher CM, the more faithful feature importance ranking.

Given the $i_{th}$ input sequence with size $n$, let $H_i$ denotes the human explanation toward feature importance ranking and $E_i$ is provided by the explanation method.

\textbf{Rank Correlation (RC)}: measure the similarity between two ranks. Spearman’s Rank Correlation Coefficient is employed to compute RC as 
\begin{equation}
 \mathrm{RC}=\frac{1}{N}\sum_{i=1}^N 1-\frac{6\sum (E_i-H_i)^2}{n_i(n_i^2-1)},
\end{equation}
where RC value ranges from $[-1,1]$.

\textbf{Overlap Rate (OR)}: measure the overlap of the top $k$ influential elements between $E_{i,k}$ and $H_{i,k}$ as 

\begin{equation}
OR(k) = \frac{1}{N}\sum \frac{1}{k}\left | E_{i,k}\cap H_{i,k} \right |.
\end{equation}

\section{Overlap rate evaluation}
\label{appendixe}
\begin{figure*}[h]
 \centering
 \begin{subfigure}[b]{0.32\textwidth}
     \centering
     \includegraphics[width=\textwidth,height=1.3in]{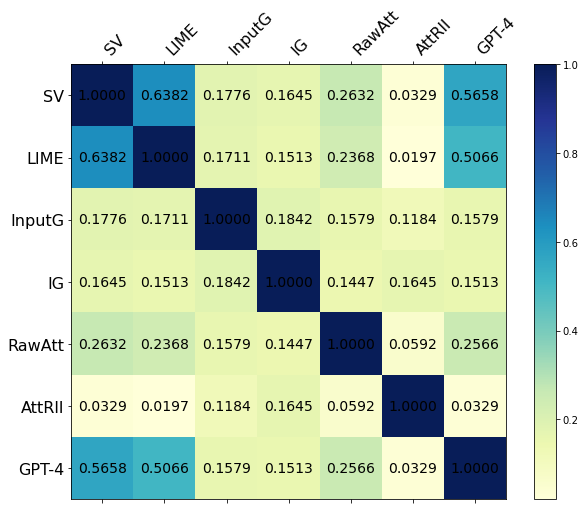}
     \caption{k = 1}
     \label{fig1a}
 \end{subfigure}
 \begin{subfigure}[b]{0.32\textwidth}
     \centering
     \includegraphics[width=\textwidth,height=1.3in]{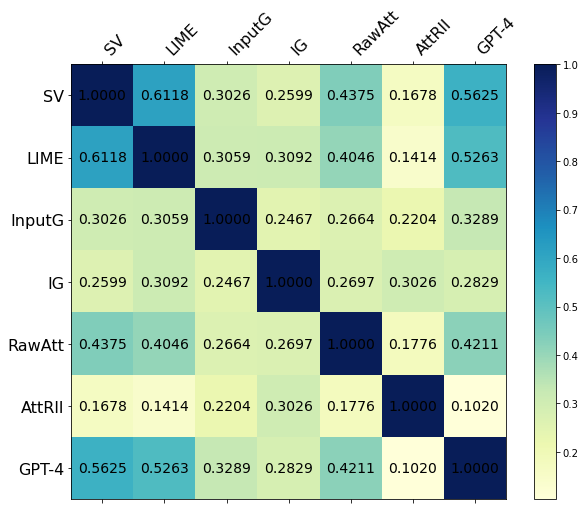}
     \caption{k = 2}
     \label{fig2a}
 \end{subfigure}
 \begin{subfigure}[b]{0.32\textwidth}
     \centering
     \includegraphics[width=\textwidth,height=1.3in]{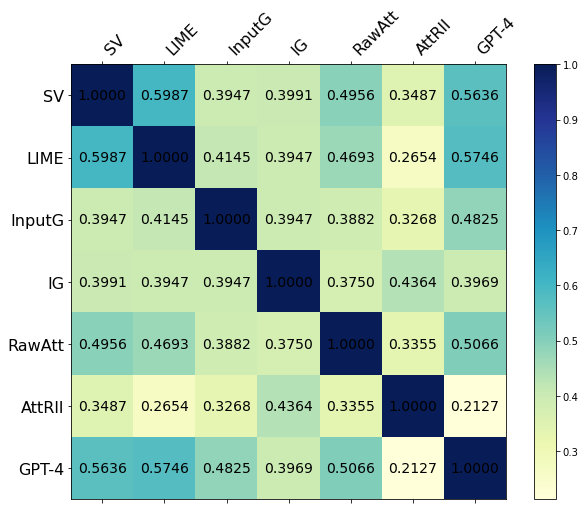}
     \caption{k = 3}
     \label{fig3a}
 \end{subfigure}
\begin{subfigure}[b]{0.32\textwidth}
     \centering
     \includegraphics[width=\textwidth,height=1.3in]{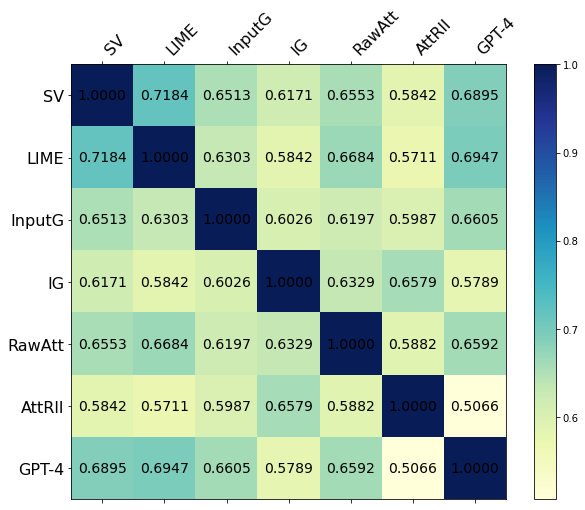}
     \caption{k = 5}
     \label{fig1b}
 \end{subfigure}
 \begin{subfigure}[b]{0.32\textwidth}
     \centering
     \includegraphics[width=\textwidth,height=1.3in]{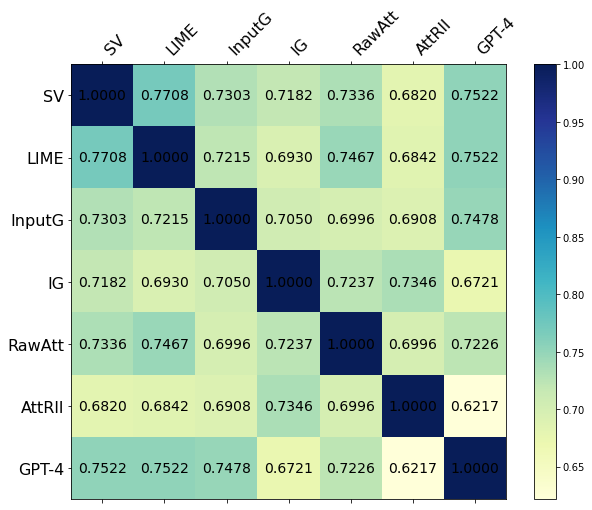}
     \caption{k = 6}
     \label{fig2b}
 \end{subfigure}
 \begin{subfigure}[b]{0.32\textwidth}
     \centering
     \includegraphics[width=\textwidth,height=1.3in]{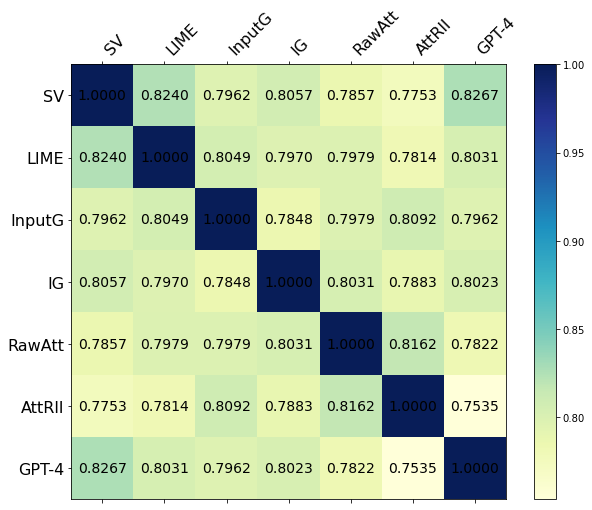}
     \caption{k=7}
     \label{fig3b}
 \end{subfigure}
\caption{Overlap rate over BERT in SST-2.}
\label{opsstb}
\end{figure*}

\begin{figure*}[h]
 \centering
 \begin{subfigure}[b]{0.32\textwidth}
     \centering
     \includegraphics[width=\textwidth,height=1.3in]{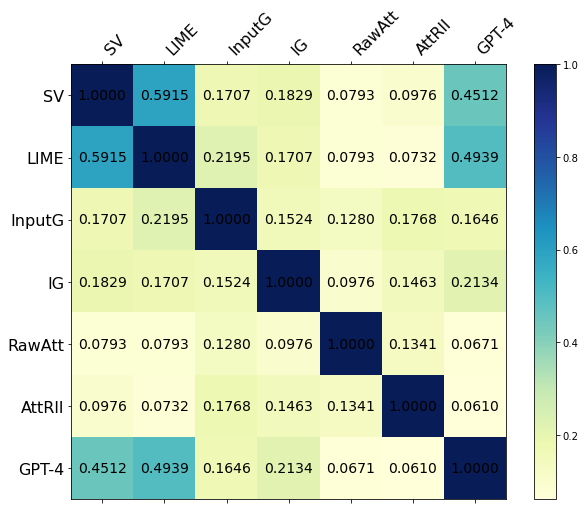}
     \caption{k = 1}
     \label{fig1t1}
 \end{subfigure}
 \begin{subfigure}[b]{0.32\textwidth}
     \centering
     \includegraphics[width=\textwidth,height=1.3in]{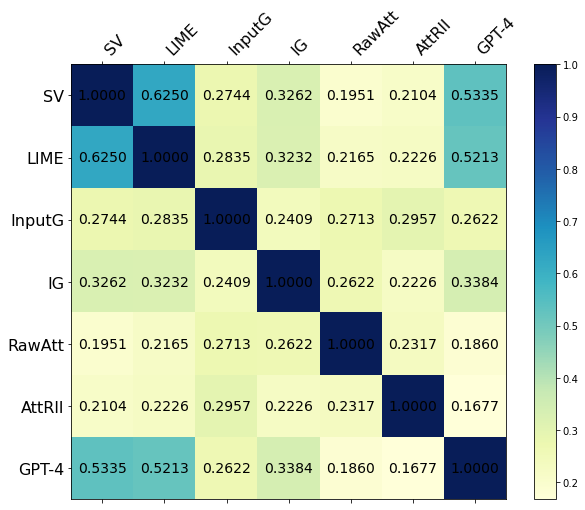}
     \caption{k = 2}
     \label{fig2t1}
 \end{subfigure}
 \begin{subfigure}[b]{0.32\textwidth}
     \centering
     \includegraphics[width=\textwidth,height=1.3in]{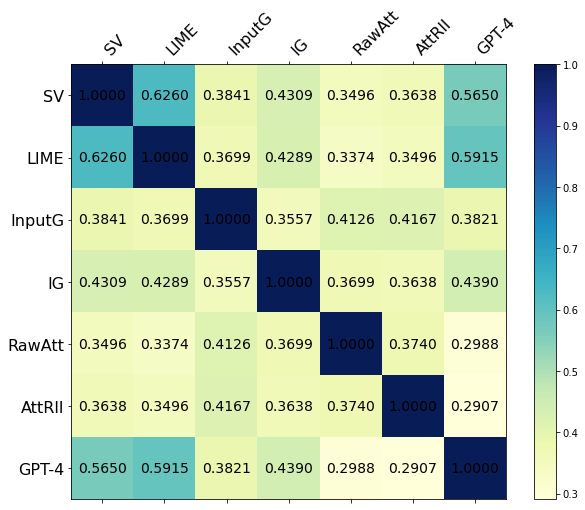}
     \caption{k = 3}
     \label{fig3t1}
 \end{subfigure}
\begin{subfigure}[b]{0.32\textwidth}
     \centering
     \includegraphics[width=\textwidth,height=1.3in]{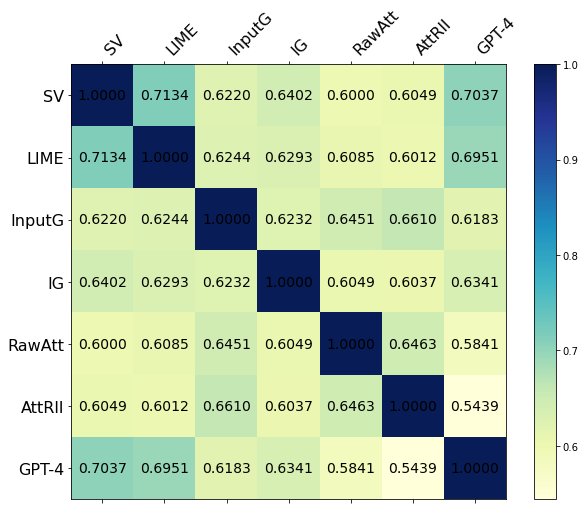}
     \caption{k = 5}
     \label{fig1t2}
 \end{subfigure}
 \begin{subfigure}[b]{0.32\textwidth}
     \centering
     \includegraphics[width=\textwidth,height=1.3in]{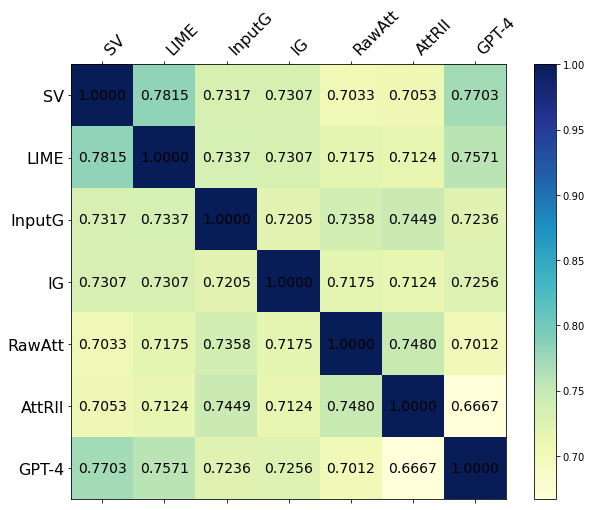}
     \caption{k = 6}
     \label{fig2t2}
 \end{subfigure}
 \begin{subfigure}[b]{0.32\textwidth}
     \centering
     \includegraphics[width=\textwidth,height=1.3in]{fig/sst_roberta_op7.png}
     \caption{k=7}
     \label{fig3t2}
 \end{subfigure}
\caption{Overlap rate over RoBERTa in SST-2.}
\label{opsstr}
\end{figure*}

\begin{figure*}[h]
 \centering
 \begin{subfigure}[b]{0.32\textwidth}
     \centering
     \includegraphics[width=\textwidth,height=1.3in]{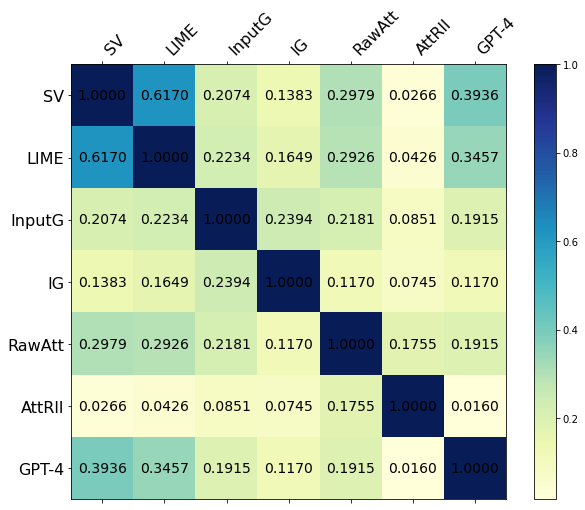}
     \caption{k = 1}
     \label{fig1z1}
 \end{subfigure}
 \begin{subfigure}[b]{0.32\textwidth}
     \centering
     \includegraphics[width=\textwidth,height=1.3in]{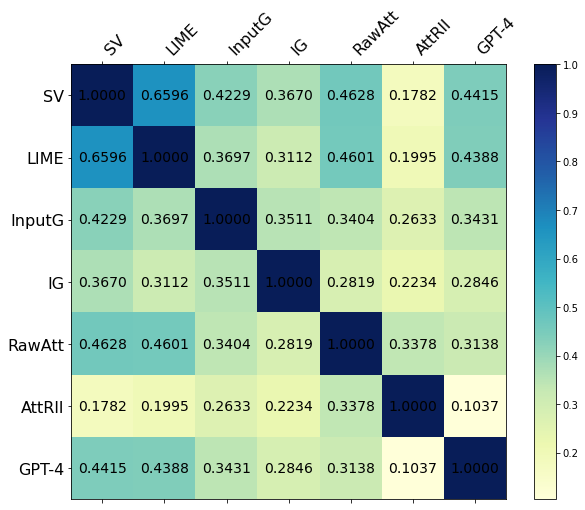}
     \caption{k = 2}
     \label{fig2z1}
 \end{subfigure}
 \begin{subfigure}[b]{0.32\textwidth}
     \centering
     \includegraphics[width=\textwidth,height=1.3in]{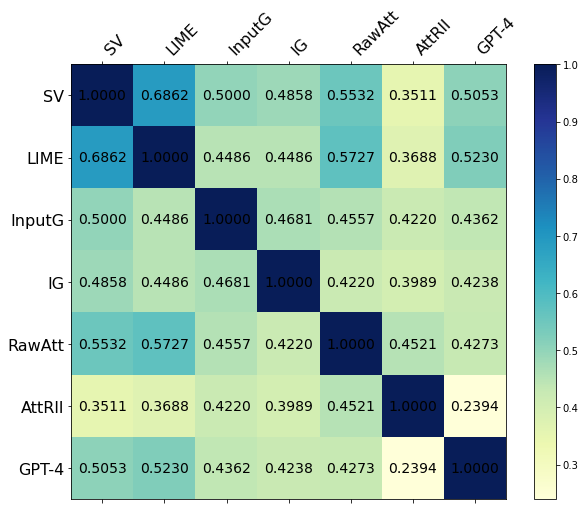}
     \caption{k = 3}
     \label{fig3z1}
 \end{subfigure}
\begin{subfigure}[b]{0.32\textwidth}
     \centering
     \includegraphics[width=\textwidth,height=1.3in]{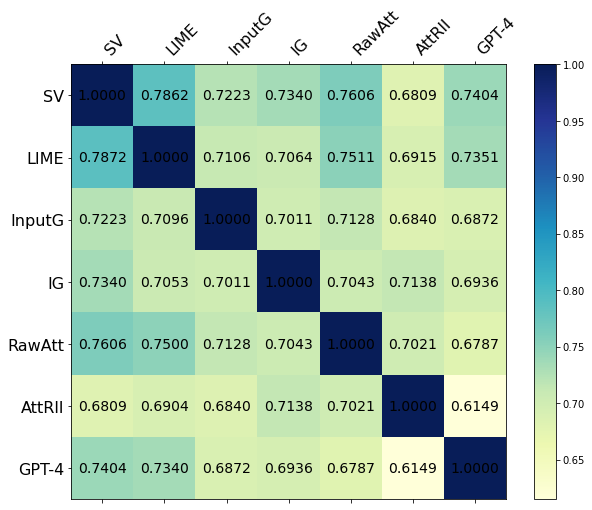}
     \caption{k = 5}
     \label{fig1z2}
 \end{subfigure}
 \begin{subfigure}[b]{0.32\textwidth}
     \centering
     \includegraphics[width=\textwidth,height=1.3in]{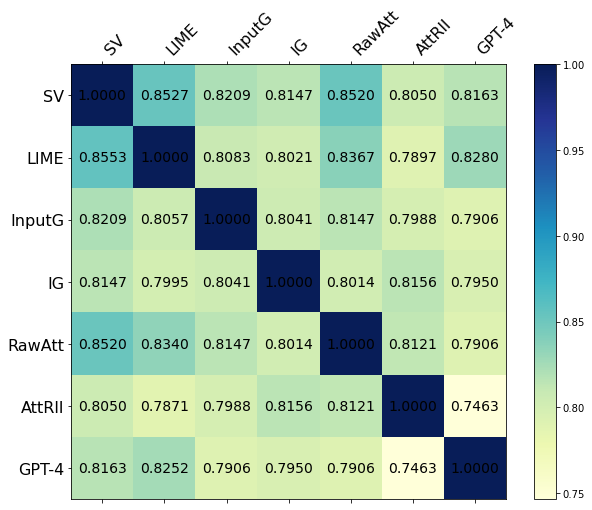}
     \caption{k = 6}
     \label{fig2z2}
 \end{subfigure}
 \begin{subfigure}[b]{0.32\textwidth}
     \centering
     \includegraphics[width=\textwidth,height=1.3in]{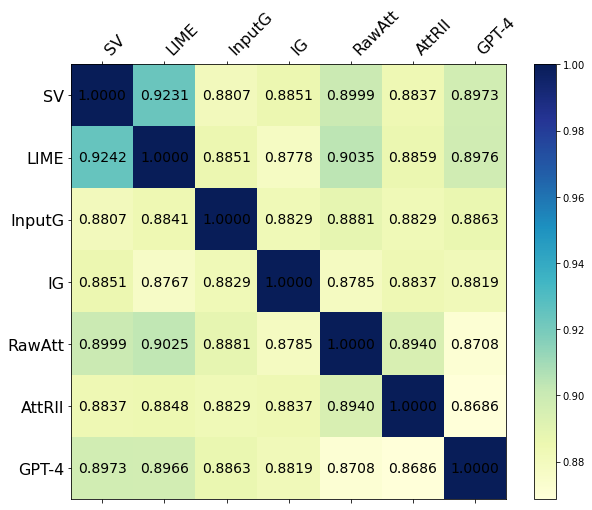}
     \caption{k=7}
     \label{fig3z2}
 \end{subfigure}
\caption{Overlap rate over BERT in SNIPS.}
\label{opsnipsb}
\end{figure*}

\begin{figure*}[h]
 \centering
 \begin{subfigure}[b]{0.32\textwidth}
     \centering
     \includegraphics[width=\textwidth,height=1.3in]{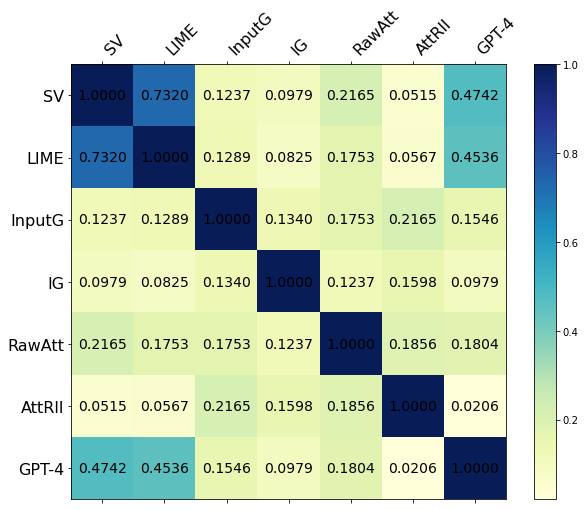}
     \caption{k = 1}
     \label{fig1k1}
 \end{subfigure}
 \begin{subfigure}[b]{0.32\textwidth}
     \centering
     \includegraphics[width=\textwidth,height=1.3in]{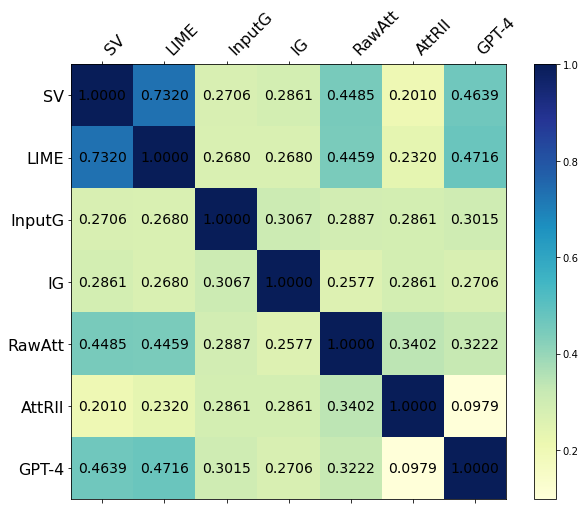}
     \caption{k = 2}
     \label{fig2k1}
 \end{subfigure}
 \begin{subfigure}[b]{0.32\textwidth}
     \centering
     \includegraphics[width=\textwidth,height=1.3in]{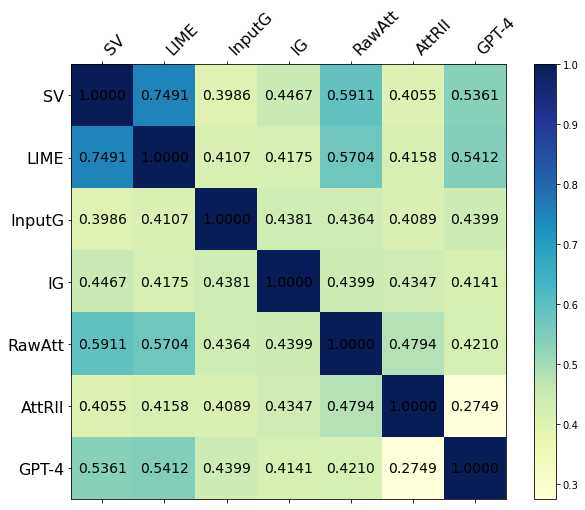}
     \caption{k = 3}
     \label{fig3k1}
 \end{subfigure}
\begin{subfigure}[b]{0.32\textwidth}
     \centering
     \includegraphics[width=\textwidth,height=1.3in]{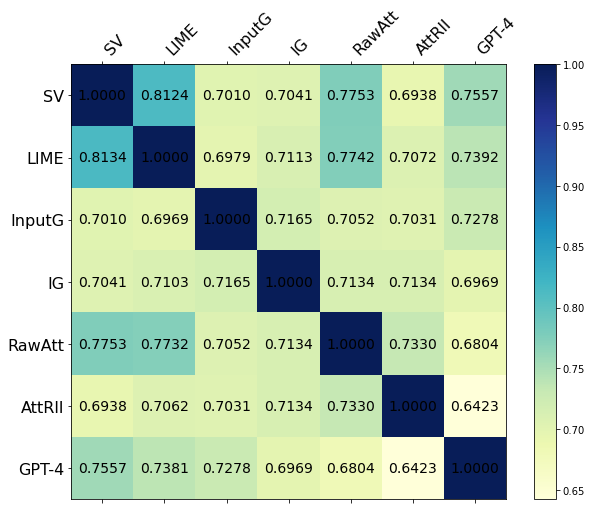}
     \caption{k = 5}
     \label{fig1k2}
 \end{subfigure}
 \begin{subfigure}[b]{0.32\textwidth}
     \centering
     \includegraphics[width=\textwidth,height=1.3in]{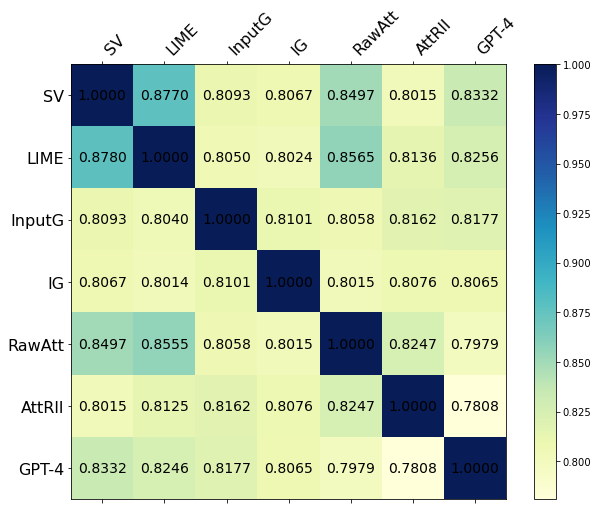}
     \caption{k = 6}
     \label{fig2k2}
 \end{subfigure}
 \begin{subfigure}[b]{0.32\textwidth}
     \centering
     \includegraphics[width=\textwidth,height=1.3in]{fig/snips_roberta_op7.png}
     \caption{k=7}
     \label{fig3k2}
 \end{subfigure}
\caption{Overlap rate over RoBERTa in SNIPS.}
\label{opsnipsr}
\end{figure*}

Fig.\ref{opsstb}, \ref{opsstr}, \ref{opsnipsb} and \ref{opsnipsr} demonstrate the overlap rate with different $k$ over BERT and RoBERTa architectures. By varying $k$ SV and LIME always have significant overlap with GPT-4.

\begin{figure*}[h]
 \centering
 \begin{subfigure}[b]{0.32\textwidth}
     \centering
     \includegraphics[width=\textwidth,height=1.3in]{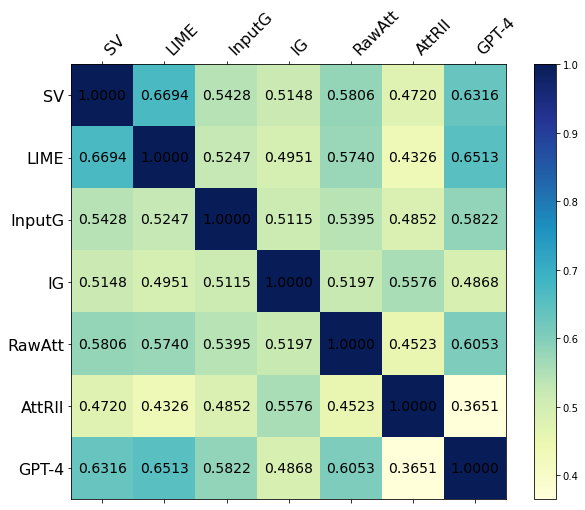}
     \caption{k = 4 over BERT on SST-2.}
     \label{fig11}
 \end{subfigure}
 \begin{subfigure}[b]{0.32\textwidth}
     \centering
     \includegraphics[width=\textwidth,height=1.3in]{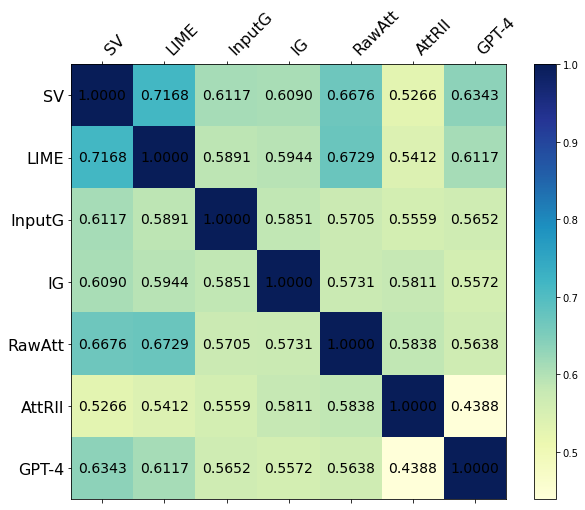}
     \caption{k = 4 over BERT on SNIPS.}
     \label{fig21}
 \end{subfigure}
 \begin{subfigure}[b]{0.32\textwidth}
     \centering
     \includegraphics[width=\textwidth,height=1.3in]{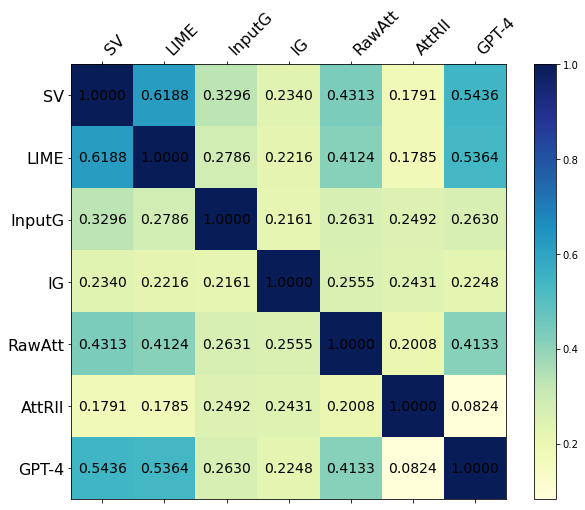}
     \caption{Over BERT on 20Newsgroup.}
     \label{fig31}
 \end{subfigure}
\begin{subfigure}[b]{0.32\textwidth}
     \centering
     \includegraphics[width=\textwidth,height=1.3in]{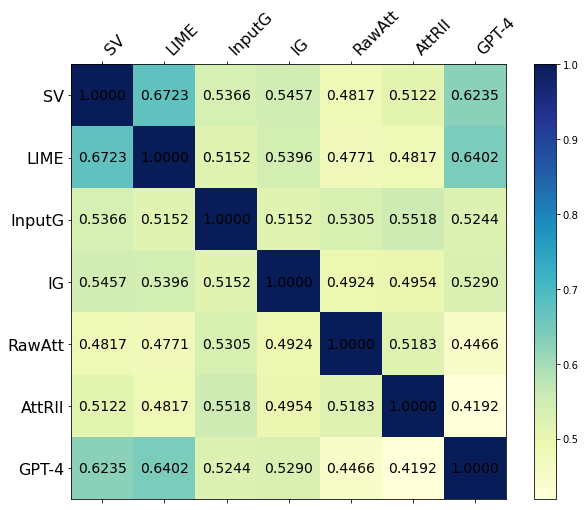}
     \caption{k = 4 over RoBERTa on SST-2.}
     \label{fig12}
 \end{subfigure}
 \begin{subfigure}[b]{0.32\textwidth}
     \centering
     \includegraphics[width=\textwidth,height=1.3in]{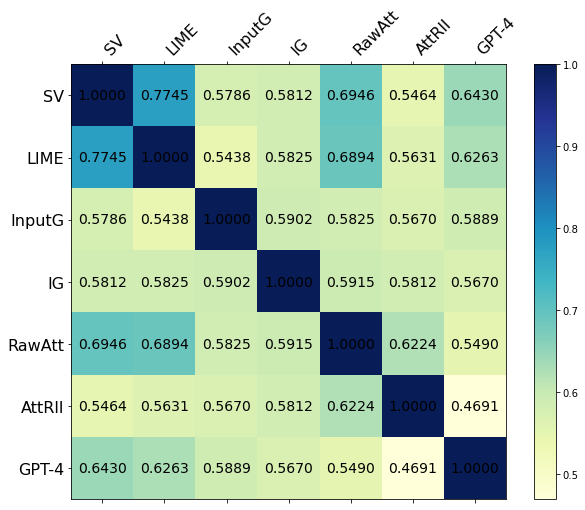}
     \caption{k = 4 over RoBERTa on SNIPS.}
     \label{fig22}
 \end{subfigure}
 \begin{subfigure}[b]{0.32\textwidth}
     \centering
     \includegraphics[width=\textwidth,height=1.3in]{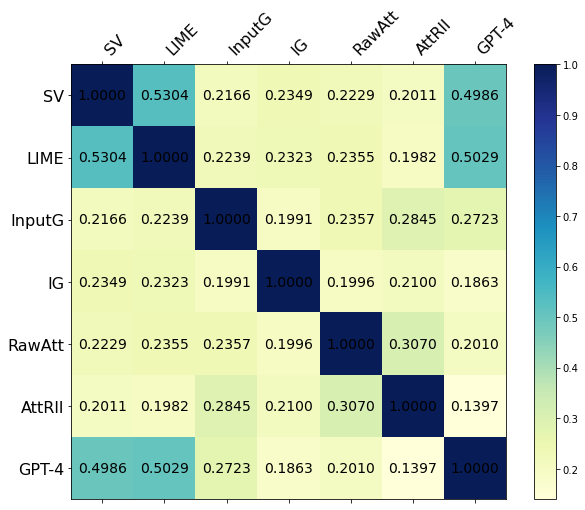}
     \caption{Over RoBERTa on 20Newsgroup.}
     \label{fig32}
 \end{subfigure}
\caption{Overlap rate when $k=4$ over BERT and RoBERTa architectures.}
\label{op4}
\end{figure*}

\end{document}